\documentclass[letterpaper]{article} 
\usepackage{aaai2026}
\usepackage{times}  
\usepackage{helvet}  
\usepackage{courier}  
\usepackage[hyphens]{url}  
\usepackage{graphicx} 
\usepackage{natbib}  
\usepackage{caption} 
\usepackage{booktabs}
\usepackage{placeins}

\usepackage{tablefootnote}
\newcommand{\mpsBox}[1]{\vskip2pt\noindent\framebox[\linewidth]{\parbox{0.97\linewidth}{#1}}\smallskip}

\usepackage{enumitem}

\usepackage{subcaption}
\usepackage{soul,xcolor,tcolorbox}
\usepackage{multirow}

\usepackage{xspace}

\newcommand{\corpus}{\textsc{TWeddit}\xspace}
\newcommand{\answerTODO}[1]{\textbf{#1}}

\usepackage[np,newcolumntypes]{numprint}
\usepackage{algorithm}
\usepackage{algorithmic}

\usepackage{newfloat}
\usepackage{listings}
\DeclareCaptionStyle{ruled}{labelfont=normalfont,labelsep=colon,strut=off} 
\floatstyle{ruled}
\newfloat{listing}{tb}{lst}{}
\floatname{listing}{Listing}
\newcommand{\hlc}[2][yellow]{{%
    \colorlet{foo}{#1}%
    \sethlcolor{foo}\hl{#2}}%
}
\definecolor{PregnancyColor}{RGB}{255,255,153}     
\definecolor{MedicalColor}{RGB}{255,204,153}       
\definecolor{MentalHealthColor}{RGB}{204,229,255}  
\definecolor{AbuseColor}{RGB}{255,153,153}         
\definecolor{SexualColor}{RGB}{255,204,229}        
\definecolor{WhiteColor}{RGB}{255,255,255}         

\title{\corpus : A Dataset of Triggering Stories Predominantly Shared by Women on Reddit}

\author{
Shirlene Rose Bandela\equalcontrib\textsuperscript{\rm 1},
Sanjeev Parthasarathy\equalcontrib\textsuperscript{\rm 2},\\
Vaibhav Garg\textsuperscript{\rm 1}
}

\affiliations{
\textsuperscript{\rm 1}Department of Computer Science, Virginia Tech, Alexandria, USA\\
\textsuperscript{\rm 2}Medlaunch Concepts, USA\\
\{shirleneroseb, vaibhavg\}@vt.edu\\
sanjeevparthasarathy@medlaunchconcepts.com
}

\begin{document}

\maketitle

\begin{abstract}

\textit{Warning: This paper may contain examples and topics that may be disturbing to some readers, especially survivors of miscarriage and sexual violence.} People affected by abortion, miscarriage, or sexual violence often share their experiences on social media to express emotions and seek support. On public platforms like Reddit, where users can post long, detailed narratives (up to 40,000 characters), readers may be exposed to distressing content. Although Reddit allows manual trigger warnings, many users omit them due to limited awareness or uncertainty about which categories apply. There is scarcity of datasets on Reddit stories labeled for triggering experiences. We propose a curated Reddit dataset, \corpus, covering triggering experiences related to issues majorly faced by women. Our linguistic analyses show that annotated stories in \corpus express distinct topics and moral foundations, making the dataset useful for a wide range of future research.
\end{abstract}


\section{Introduction}

According to surveys, 1 in 6 women in the United States has experienced extreme form of sexual violence such as rape \cite{RAINN-25:sexual-assault}. In addition, 15\% women who are pregnant suffer at least one miscarriage in their lifetime \cite{MiscarriageStatistics2015}. Some women facing such issues leverage Reddit to vent out their experiences anonymously. Reddit hosts multiple subreddits, such as \url{r/SexualHarassment} and \url{r/Miscarriage}, where \emph{posters} sharing their stories get emotional support and advice from other Redditors called \emph{helpers}.

Helpers who have gone through experiences similar to posters, might feel triggered while reading these stories and assisting them \cite{wiegmann2023trigger}. Although Reddit facilitates posters to add trigger warnings manually, many posters don't add them (only 5.3\% posters according to our analysis), which may be mentally deteriorating for helpers. While majority of existing research focuses on analyzing malicious speech (hate, fear, and inciting speech), there is a scarcity of work attending to non-malicious yet emotionally triggering content.  \citet{wiegmann2023trigger} generated trigger warnings on fanfiction stories to alert readers. However, their focus was on fanfiction stories, not Reddit. The nature of language in fanfiction stories is different from Reddit stories (shown through F1 score below). We focus on trigger warnings on Reddit stories describing two major issues faced by women: (i) stories highlighting pregnancy concerns (including miscarriages and abortion) and (ii) stories highlighting sexual violence. The victims of sexual violence involve other genders as well (not just women), however the stories in \corpus are predominantly written by women (85.4\% of posters are women as described in stories). This also conforms to the gender-based statistics in the US showing women being the major victim of sexual violence reported cases \cite{NSVR-23:metoo-statistics}.

To this end, we curate, \corpus\footnote{https://doi.org/10.5281/zenodo.18251622} \cite{Tweddit:dataset-Zenodo},
a dataset of \np{5000} Reddit stories from 22 subreddits (pertaining to above topics) labeled for seven trigger categories based on a prior taxonomy. One story can correspond to multiple trigger categories as well. Hence, it is a daunting annotation task. To do so, we leveraged semi-supervised active learning paradigm. In addition, we provide agreement scores between annotators as metrics of reliability. 



Wiegmann et al.'s model (state of the art) on \corpus yielded only 36.81\% macro F1 score, lending credence to our novelty claims. Our linguistic analysis on \corpus revealed various interesting findings. Topic modeling reveals that \textit{Pregnancy} and \textit{Medical} stories emphasize procedural and clinical terms that could be triggering (e.g., `gestational sac', `molar pregnancy', `bleeding'), while \textit{Abuse} and \textit{Mental Health} categories highlight `coercion', `distress', and `coping'. The stories across \corpus show moral patterns aligned with Moral Foundation Theory (MFT). Care, Authority, and Loyalty are the top 3 foundations. Most users describe harm, question autonomy, and talk about their fears of betrayal and broken relationships. In the future, \corpus can be used by researchers for various language processing tasks, along with the automatic generation of trigger warnings on pregnancy and sexual violence related subreddits.

\section{The Setting : Triggering Stories Require Warnings}

We present two examples including excerpts from triggering stories. Since these stories are very long due to high character limit on Reddit, we are highlighting only the main triggering sentences in them. In addition, we mention the trigger-warning category associated with each highlighted text (each category is described later in this section).

Example 1 shows an abortion experience (from r/abortion) in which the poster describes navigating a physically and emotionally challenging medical process and seeks support on how to manage the uncertainty surrounding it. Example 2 presents a violence story (from r/sexualassault) in which the poster reflects on a coercive sexual encounter and describes the lasting effects they had on their sense of safety and intimacy.
To be used as examples in this paper, we have paraphrased
these stories so that the traceability of the survivor and their details is minimal. 

\begin{tcolorbox}[colback=white, colframe=black!60, boxrule=0.5pt, arc=3pt, width=\linewidth]
\textbf{Categories:}  
\hlc[MedicalColor]{Medical}, 
\hlc[MentalHealthColor]{Mental Health}, 
\hlc[AbuseColor]{Abuse}, 
\hlc[SexualColor]{Sexual}\\[2pt]

\textbf{Example 1 : Excerpts from an abortion story}\\[4pt]
\hlc[WhiteColor]{...}
\hlc[MedicalColor]{I had four extra pills but decided to stop because I was already bleeding so much}
\hlc[WhiteColor]{...}
\hlc[MentalHealthColor]{I was advised to wait for another 2--3 weeks before taking another test, but the anxiety was killing me already.}
\hlc[WhiteColor]{...}\\[2pt]

\textbf{Example 2 : Excerpts from a sexual violence story}\\[4pt]
\hlc[AbuseColor]{I repeatedly said no, but then she forcibly fondled me until I relented.} 
\hlc[WhiteColor]{......}
\hlc[SexualColor]{Since then, about 90\% of the time I have sex, I can't actually maintain an erection long enough to have sex.}
\hlc[WhiteColor]{......}
\hlc[MentalHealthColor]{Even when I'm enjoying fooling around a part of my mind is suddenly seized with the panic and helplessness I felt that night.}\hlc[WhiteColor]{......}
\end{tcolorbox}

In the above mentioned subreddits, helpers may have gone through similar experiences and could feel triggered reading specific details mentioned in these stories \cite{wiegmann2023trigger}. Our goal is not to prevent the poster from sharing vivid descriptions on these support communities, but, rather utilize future computational approaches that can generate trigger warnings for such descriptions. In order to advance such future research, we present \corpus. 

\textbf{Dataset Description} \corpus \cite{Tweddit:dataset-Zenodo} consists of \np{5000} Reddit stories collected from approximately 22 support-oriented subreddits. In the dataset, each instance contains the story's title, the story's body, and a set of trigger-warning categories associated.

\subsection{Trigger-Warning Categories}
\label{sec:definitions}



\citet{wiegmann2023trigger} provide a taxonomy for trigger warnings, containing seven broad categories.
We manually read 500 Reddit stories and found that no triggering experience falls beyond these seven trigger-warning categories. Hence, we adopt the same taxonomy for annotation. We describe these seven categories below.

\paragraph{Pregnancy}
Applied to stories that explicitly mention pregnancy, pregnancy loss, miscarriage, abortion (including medication abortion and surgical abortion), chemical pregnancy, or general discussion of abortion or miscarriage as pregnancy-related outcomes. 


\paragraph{Medical}
Applied to stories containing concrete medical details tied to abortion, miscarriage, or sexual violence, such as pills, dilation and curettage (D\&C), HCG levels, procedures, ultrasounds, emergency room visits, antibiotics, confirmed heavy bleeding, or clinical examinations. 

\paragraph{Mental Health}
Applied to stories expressing explicit or clearly observable emotional or psychological distress caused by the core topic (pregnancy issues and sexual violence). This included panic attacks, severe anxiety, intrusive thoughts, grief, guilt, sadness, crying, or being emotionally overwhelmed. 


\paragraph{Abuse}
Applied to stories describing coercion, control, threats, or sexual assault by another person. This included pressuring someone about abortion, rape, stalking, intimidation, isolation, threats to report or damage reputation, or sexual assault through force, coercion, or exploitation of vulnerability. 


\paragraph{Aggression}
Applied to stories describing non-sexual physical violence, such as hitting, slapping, choking, restraining, or throwing objects, where there was clear evidence of physical harm. Verbal aggression alone did not qualify for this category.

\paragraph{Sexual}
Applied to stories containing references to consensual sexual content, behavior, or sexual acts (e.g., different types of sex, pornography), as well as sexual health topics such as contraception or libido. 


\paragraph{Discrimination}
Applied to stories describing stigma, shaming, moral policing, mistreatment in clinical settings, denial of care, harassment, or derogatory remarks applied directly to the individual. 



\textbf{Organization} We describe how we collected Reddit stories in Section 3. Section 4 describes the curation of \corpus using an active learning framework. In Section 5, we evaluate multiple models for trigger-warning prediction task. Section 6 presents exploratory analyses of \corpus, including topic modeling, moral foundations, and so on. We list related work in Section 7. Section 8 concludes the paper and provides directions for future work.

\section{Collecting Reddit Stories}
\citet{garg2025analyzing} curated a dataset of \np{4934} sexual violence stories posted on three subreddits, r/meToo, r/SexualHarassment, and r/sexualassault. Since it's a recent dataset, we leveraged it for our study. For pregnancy-related concerns, we first identified \np{22} subreddits that feature such experiences. Previous studies \cite{stanier2024polarization} list bigrams and trigrams related to traumatic experiences during or after pregnancies. We leveraged these keywords to filter relevant stories while scraping through Python's Reddit API Wrapper (PRAW). In total, we collected \np{6431} pregnancy-related stories. To check the relevance of keywords, we randomly sampled \np{100} of \np{6431} stories and found \np{91} of them are relevant to pregnancy-related concerns. 
To sum up, we collected \np{6431} pregnancy-related stories (\np{4578} abortion-related and \np{1871} miscarriage-related) using PRAW and \np{4934} sexual violence stories from the prior work \cite{garg2025analyzing}.

\begin{table}[h!]
\centering
\caption{Cohen’s Kappa scores for three rounds. According to these values, the Kappa scores varied between moderate and substantial agreement.\tablefootnote{For Aggression in round 1, the sample size was very small (only 2 stories)}}
\label{tab:kappa_scores}
\begin{tabular}{lccc}
\toprule
\textbf{Tag} & \textbf{Round 1} & \textbf{Round 2} & \textbf{Round 3} \\
\midrule
Abuse         & 0.9768 & 0.9561 & 0.9231 \\
Aggression    & ----- & 0.7899 & 0.6575 \\
Discrimination& 1.0000 & 0.8837 & 0.7839 \\
Medical       & 0.8792 & 0.8160 & 0.8408 \\
Mental Health & 0.7883 & 0.7317 & 0.7915 \\
Not Applicable& 0.6622 & 0.6622 & 0.9280 \\
Pregnancy     & 0.9787 & 0.9566 & 1.0000 \\
Sexual        & 0.9597 & 0.7340 & 0.9170 \\
\bottomrule
\end{tabular}
\end{table}

\section{Curating \corpus Corpus}
\begin{figure}[t]
\centering
\includegraphics[width=\columnwidth]{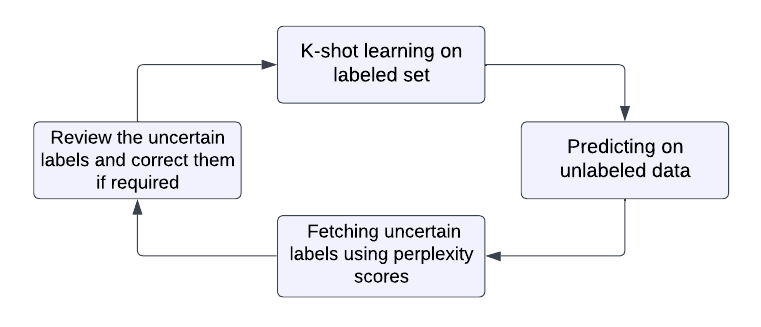}
    \caption{Active learning loop where model predictions on unlabeled data are filtered using perplexity-based uncertainty, reviewed, corrected, and added back to the labeled set for iterative k-shot learning.}

    \label{fig:alcycle}
\end{figure}

Since manual annotation is expensive and time-consuming, LLMs have been extensively utilized for labeling large datasets \cite{zong2025empowering}. Moreover, in our case, recruiting annotators for reading disturbing stories would be an ethical concern. As a result, we adopted active learning strategy (uncertainty sampling) involving LLMs. As shown in Figure~\ref{fig:alcycle}, our active learning cycle contained four steps.  First, we manually labeled an initial set of stories for trigger warnings and tried multiple LLMs using K-shot learning (Section~\ref{sec:fewshot}). Second, using this in-context learning, we asked LLMs to predict trigger categories on a pool of unlabeled stories (Section~\ref{sec:unlabeled}). Some predictions would be wrong. Hence, third, we sampled uncertain labels using perplexity scores. Fourth, we manually reviewed those uncertain labels and corrected them if necessary (Section~\ref{sec:unlabeled}). The new labeled stories were added to the already annotated set. This cycle was repeated multiple times to produce \corpus (Section~\ref{sec:additional}).   

Using taxonomy suggested by \citet{wiegmann2023trigger}, each Reddit post was labeled for seven trigger categories: \emph{Abuse}, \emph{Discrimination}, \emph{Aggression}, \emph{Medical}, \emph{Pregnancy}, \emph{Sexual}, and \emph{Mental Heath}.


\subsection{K-Shot Learning on Initial Set}
\label{sec:fewshot}
\subsubsection{Labeling Initial Set of Stories} 
\label{sec:InitialData}
The first two authors of this paper labeled 500 stories over five rounds. For labeling each story, annotators' instructions included following steps.

\begin{itemize}
    \item Read the given story along with its title to be annotated 
    \item Review the definitions for each trigger category (provided in Section~\ref{sec:definitions})

    \item For each of the categories,  label `YES' if the story included any descriptions pertaining to that category. Otherwise, label `NO'

\end{itemize}

\textbf{Data Quality} After each round, they resolved disagreements. We obtained moderate to substantial agreement for all seven categories as shown in Table~\ref{tab:kappa_scores}. Leading datasets \cite{Hatexplain2021} also achieve comparable agreement scores.

\subsubsection{Prompt Construction for K-Shot Learning}
We leveraged GPT-4o-mini (OpenAI) and LLaMA-3-8B for in-context learning on the initial labeled set. For K-shot learning, we provided K labeled examples (from initial set) along with the definitions of each trigger category. After multiple attempts, we finalized our prompt, which is shown below (lightly edited for grammar). 


{
\mpsBox{
``You are an attentive content reviewer identifying specific emotional or traumatic triggers in Reddit posts related to Abortion (A), Miscarriage (M), or Sexual Violence (SV). If the post is NOT clearly A, M, or SV, then mark it as 'Not Applicable' and STOP. Your task is to analyze the post and decide whether readers could be triggered by it. We have multiple categories of trigger warnings, but in this case, you will evaluate only one trigger label at a time using a careful step-by-step approach. First, carefully read the post (Title and Description). Second, compare the content to \textless category-name\textgreater
. Third, use the few-shot examples and their labels, but judge THIS category only. Fourth, decide if the post explicitly and clearly matches this category. Fifth, if it does then answer YES; otherwise, NO.  

\emph{Each category is then explained using its definition}
}
}

\citet{Prompt-Engineering2025} suggested a few ways for effective prompting. We employed them to design our prompt. For instance, we asked LLMs to predict categorical labels (YES or NO) instead of annotating on a Likert scale of ratings. We set the temperature of LLMs to zero to mitigate any randomness in predictions. As suggested, we focused on providing clear definitions of each of the trigger categories. Moreover, we leveraged the chain-of-thought prompting strategy to help LLMs understand the task well and predict.

We tested both the LLMs for K ranging from 1 to 5 as seen in Table~\ref{tab:kshot_comparison}. For $K=3$ of GPT-4o-mini, we achieved the best macro F1 of 78.3\%. Hence, we utilized GPT-4o-mini for prediction tasks in each active learning cycle.






\subsubsection{Cosine Similarity}
The performance of LLMs depends on the examples that are shown to them \cite{LLM-ActiveLearning2023}. Recent studies have started using `similarity' as an important criteria in choosing K examples \citep{LLM-ActiveLearning2023}. Hence, we selected K examples using cosine similarity. For each input post, we computed its embedding (using MPNET-based sentence-transformer \cite{MPNET-SentenceTransformer}) and retrieved the K (K is 3) most similar examples from our labeled set. This approach selects similar examples that are contextually aligned with the input, providing effective context for GPT-4o-mini to predict. 

\subsection{Predicting on Unlabeled Data and Fetching Uncertain Predictions}
\label{sec:unlabeled}


In each active learning cycle, we sampled \np{500} stories that need to be annotated. We leveraged the few-shot setup (described in Section~\ref{sec:fewshot}) to predict trigger categories on these 500 stories. However, some predictions from GPT-4o-mini could be wrong. Hence, we fetched its uncertain predictions using the perplexity metric and corrected those labels if needed. Perplexity metric assesses how confident the LLM is while making such predictions. The higher the perplexity score, the more chances of that prediction being wrong, making perplexity a good candidate for fetching uncertain predictions. For each of the \np{500} stories, we recorded the maximum perplexity value obtained over predicting seven trigger categories. 




\subsubsection{Threshold on Perplexity Scores}
For first four active learning cycles, we also manually annotated \np{500} sampled stories for trigger categories. This ground truth helped us to find a threshold on perplexity scores such that the mispredicted stories (according to the ground truth) fall above that threshold. In that case, for correcting the mispredicted labels, we had to only review the predictions having perplexity higher than the threshold. In other words, the threshold serves as a way of uncertainty sampling, as practiced in active learning paradigm.
To calculate the threshold, we followed a prior study \cite{garg2025analyzing} applying this paradigm. We found 1.0007 as the perplexity threshold based on achieving high recall (while also maintaining a good precision value) at identifying mispredicted cases for four consecutive rounds. For these rounds, we present scores and additional details in appendix (Tables~\ref{tab:ppl_threshold_r1}--\ref{tab:ppl_threshold_r4}). We support this threshold by also presenting ROC curves for all rounds, as shown in Figure~\ref{fig:roc_all_rounds} of appendix.

In each round, out of 500 stories, we selected the stories having perplexity scores above the threshold for verification. The first two authors manually verified the selected stories and corrected their labels in case of mispredictions. This new labeled data (500 stories) was added to the initial labeled data (500 stories described in Section~\ref{sec:InitialData}). That means, after the first cycle, our dataset contained \np{1000} labeled stories.



\subsection{Additional Details}
\label{sec:additional}

We repeated the above steps for 9 cycles. After our threshold was fine-tuned using first five rounds, we applied it to filter and correct uncertain predictions in the rest of the four cycles. After all cycles, this paradigm produced \corpus, which contains \np{5000} stories labeled for seven trigger categories.

\begin{table}[t]
\centering
\small
\caption{Performance comparison of GPT and LLaMA across k-shot configurations. Best F1 for each model is in bold.}
\label{tab:kshot_comparison}
\begin{tabular}{c|ccc|ccc}
\toprule
\multirow{1}{*}{$k$} &
\multicolumn{3}{c|}{\textbf{GPT}} &
\multicolumn{3}{c}{\textbf{LLaMA}} \\
& P & R & F1 & P & R & F1 \\
\midrule
1 & 0.812 & 0.793 & 0.778 & 0.823 & 0.700 & 0.714 \\
2 & 0.815 & 0.781 & 0.778 & \textbf{0.826} & \textbf{0.711} & \textbf{0.722} \\
3 & \textbf{0.832} & \textbf{0.784} & \textbf{0.783} & 0.828 & 0.700 & 0.716 \\
4 & 0.776 & 0.780 & 0.755 & 0.823 & 0.706 & 0.718 \\
5 & 0.823 & 0.775 & 0.772 & 0.823 & 0.706 & 0.718 \\
\bottomrule
\end{tabular}

\end{table}

\begin{table*}[t]
\centering
\small
\begin{tabular}{lcccccc}
\toprule
\textbf{Model} &
\textbf{Macro-F1} &
\textbf{Macro-Precision} &
\textbf{Macro-Recall} \\
\midrule
\cite{wiegmann2023trigger}'s longformer & 0.368 & 0.571 & 0.322 \\
\midrule
TF-IDF + SVM     & 0.670 & \textbf{0.826} & 0.647 \\
TF-IDF + LR      & 0.682 & 0.665 & 0.795 \\
GloVe + SVM      & 0.654 & 0.594 & 0.818 \\
GloVe + LR       & 0.639 & 0.576 & 0.813 \\
Word2Vec + SVM   & 0.659 & 0.598 & 0.828 \\
Word2Vec + LR    & 0.645 & 0.579 & 0.827 \\
\midrule
BERT             & 0.729 & 0.684 & \textbf{0.836} \\
RoBERTa          & 0.740 & 0.706 & 0.803 \\
XLNet            & \textbf{0.763} & 0.752 & 0.779 \\
\bottomrule
\end{tabular}
\caption{Performance comparison of traditional and transformer models for multi-label trigger-warning classification on \corpus. The highest performance in each column is shown in bold.}
\label{tab:model_comparison}
\end{table*}

\section{Evaluating Models for Predicting Trigger Warnings}

\citet{wiegmann2023trigger} developed a tool using the Longformer model that identifies trigger warnings from fanfiction stories. We investigated if their tool can be applied to Reddit stories. On \corpus, their tool achieved a macro F1 score of 36.81\% (57.1\% macro-precision and 32.2\% macro-recall). This concludes the tool is limited to identifying trigger warnings from only fanfiction stories, and may not generalize well to annotated Reddit stories.

Table~\ref{tab:model_comparison} compares their tool with embeddings-based and transformer-based approaches trained on \corpus. Overall, transformer-based models (BERT, RoBERTa, XLNet) consistently outperform traditional embeddings-based methods, achieving higher macro-F1 performance by better capturing contextual dependencies and semantic cues in sensitive stories. Among these transformer models, XLNet achieves the highest performance with a macro-F1 of 0.763.

However, despite the advantages of transformer architectures, macro-F1 scores remain below 0.80 across all models. This gap reflects the inherently subtle, overlapping, and context-dependent nature of emotionally triggering language, where cues are often implicit rather than explicitly stated. This underscores the importance of building future robust models for predicting trigger warnings.

\section{Exploratory Analysis of \corpus}

\subsection{Distribution of Trigger Categories}
In \corpus, we found \np{3553} stories in the \textit{Mental Health} category and \np{2653} in \textit{Pregnancy} category. This reflects posters generally share triggering experiences related to their psychological well-being, their feelings and emotions, and their reproductive health. Moreover, other prominent trigger categories include \textit{Medical} (1815 stories) and \textit{Abuse} (1706 stories), while categories like \textit{Aggression} (196 stories) and \textit{Discrimination} (526 stories) are comparatively sparse.

\begin{figure}[!h]
    \centering
    \includegraphics[width=1\linewidth]{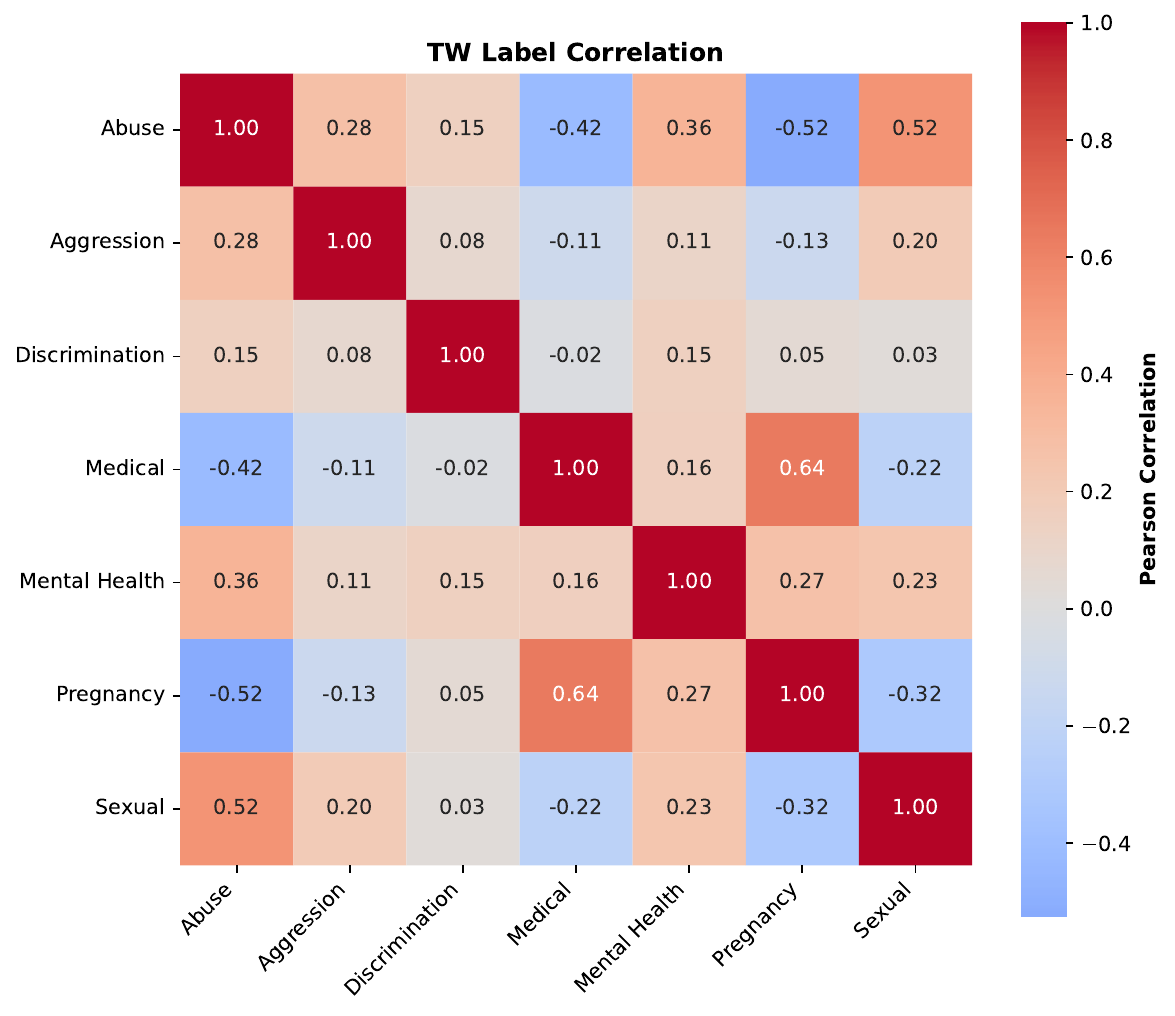}
\caption{Pearson correlation matrix across trigger warning categories, showing positive associations (Medical-Pregnancy and Abuse-Sexual) and negative associations (Pregnancy-Abuse).}
    \label{fig:Label_corr}
\end{figure}

The correlation matrix in Figure~\ref{fig:Label_corr} reveals meaningful co-occurrence patterns. We found \textit{Medical} category strongly correlating with \textit{Pregnancy}, indicating narratives describing clinical procedures around reproductive health. \textit{Abuse} and \textit{Sexual} categories show moderate positive correlation due to shared context in sexual trauma-related stories.


\begin{figure}[!h]
    \centering
    \setlength{\fboxsep}{0pt}
    \setlength{\fboxrule}{0.4pt}

    \begin{subfigure}[t]{0.30\linewidth}
        \centering
        \fbox{\includegraphics[width=2.7cm, height=3.75cm]{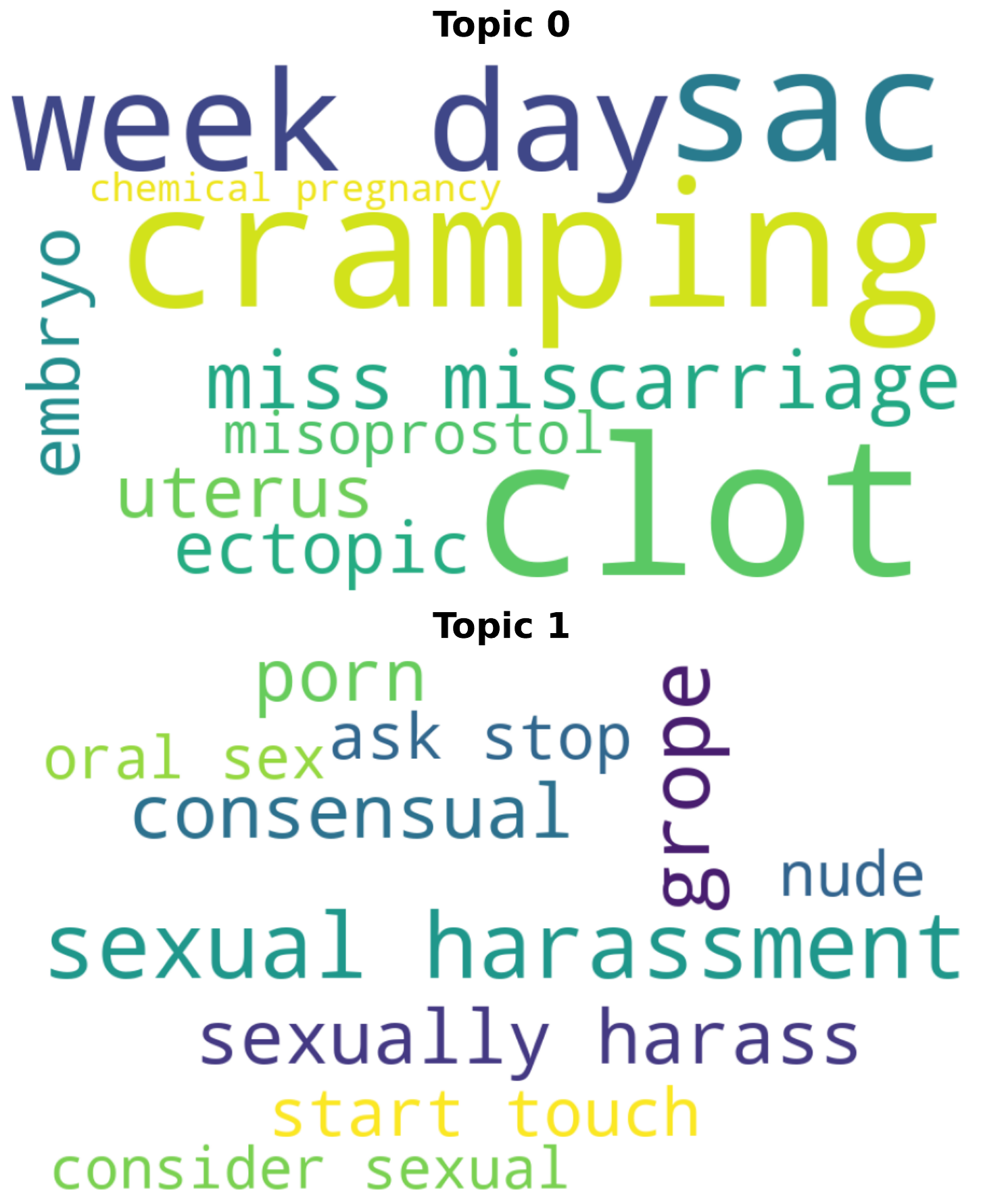}}
        \caption{Mental Health}
        \label{fig:WC_MentalHealth}
    \end{subfigure}
    \hfill
    \begin{subfigure}[t]{0.295\linewidth}
        \centering
        \fbox{\includegraphics[width=2.7cm, height=3.75cm]{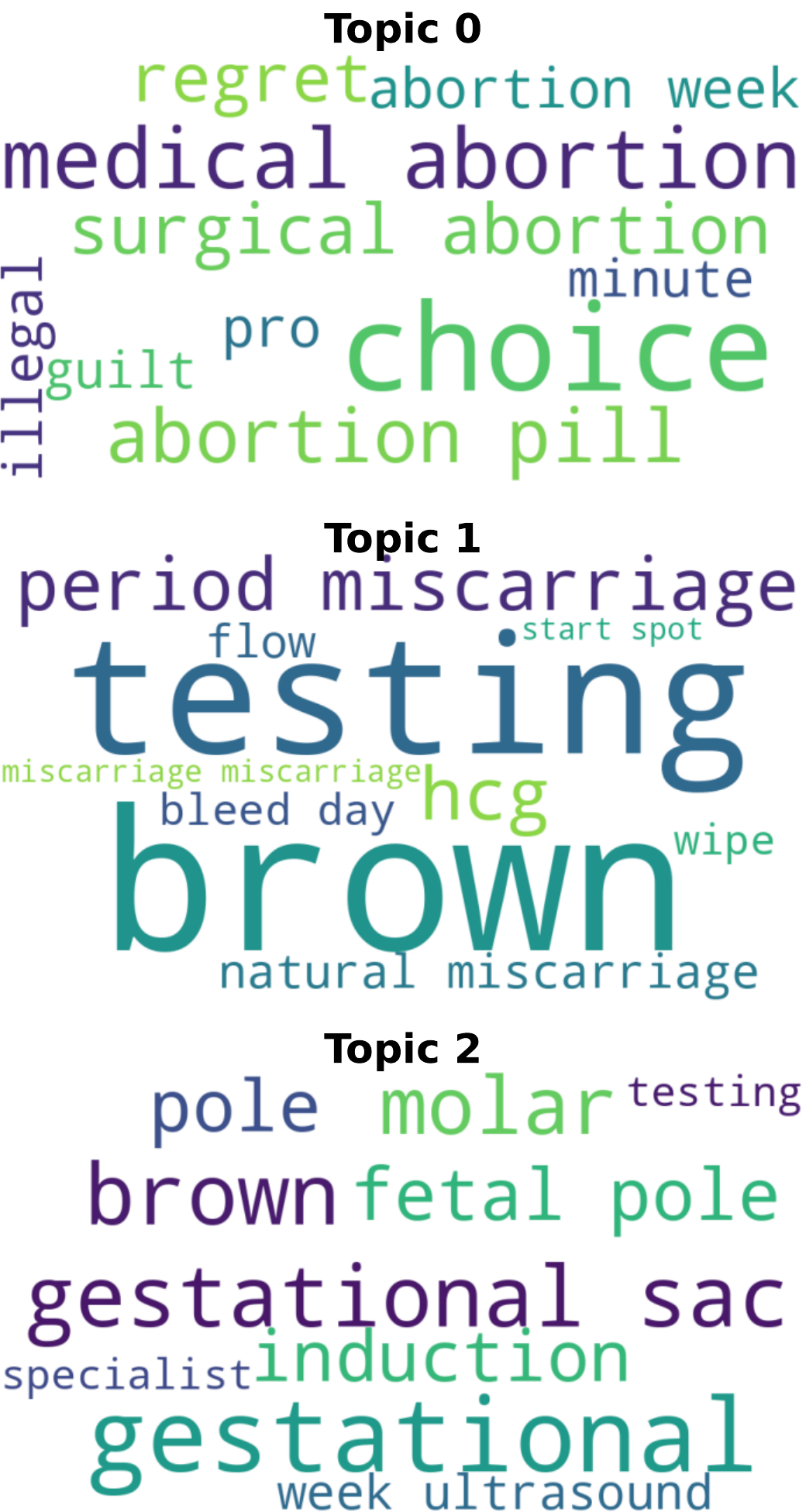}}
        \caption{Pregnancy}
        \label{fig:WC_Pregnancy}
    \end{subfigure}
    \hfill
    \begin{subfigure}[t]{0.30\linewidth}
        \centering
        \fbox{\includegraphics[width=2.7cm, height=3.75cm]{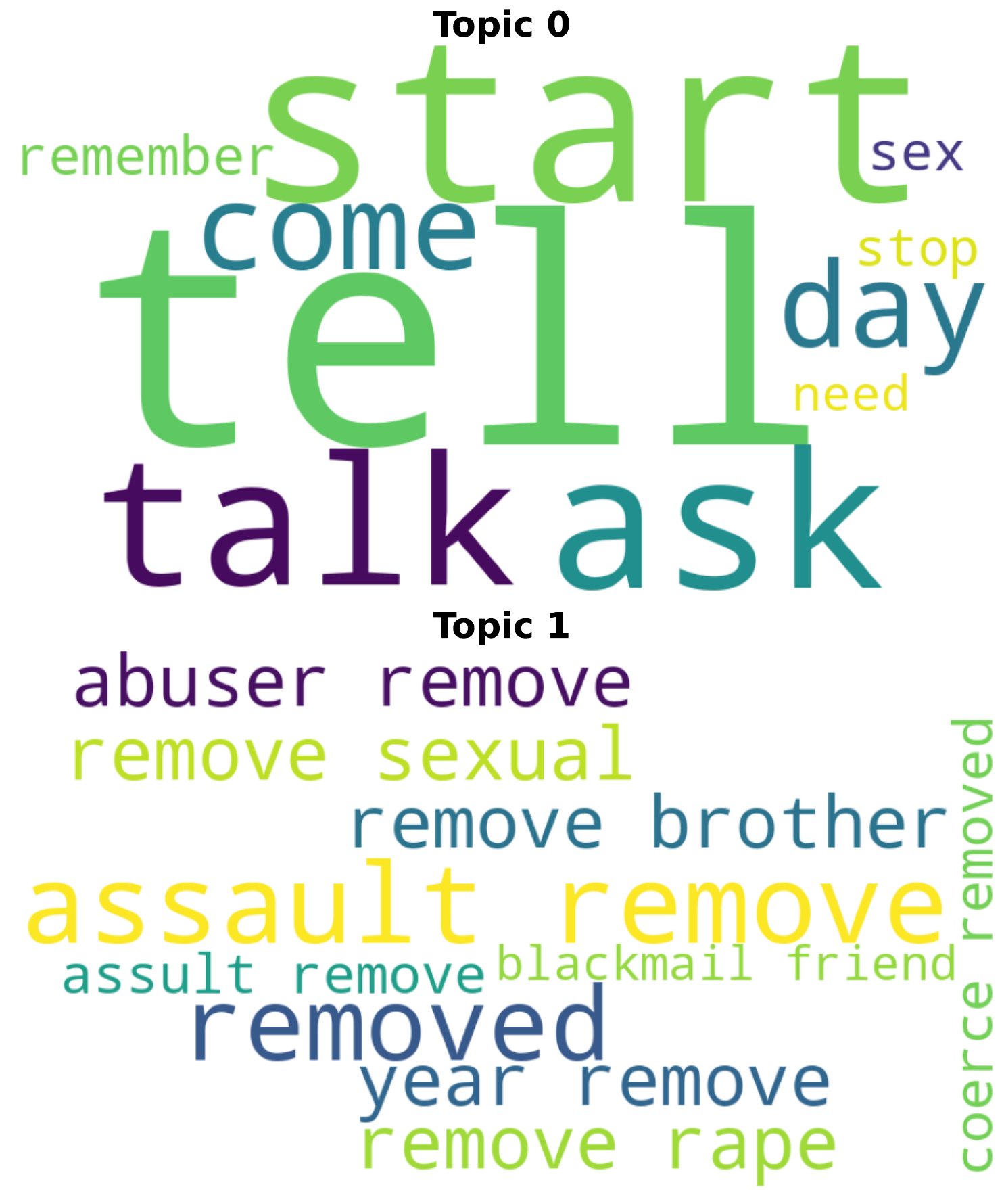}}
        \caption{Abuse}
        \label{fig:WC_Abuse}
    \end{subfigure}

\caption{Topic word clouds for three dominant trigger categories : \emph{Mental Health, Pregnancy, and Abuse}, highlighting the most salient terms in each category and illustrating distinct thematic language patterns across triggers.}

    \label{fig:topic_wordclouds}
\end{figure}

\subsection{Topic Modeling}
Using BERTopic, we conducted topic modeling on four most prominent trigger categories : \textit{Mental Health, Pregnancy, Medical and Abuse}. Since Pregnancy and Medical had overlapping topics, we only discuss one of both here. Among the stories labeled as \textit{\textbf{Mental Health}} category, one topic revolves around pregnancy complications and miscarriage (shown in upper half of Figure~\ref{fig:WC_MentalHealth}), with terms like \textbf{\textit{clot}, \textit{cramping}, and \textit{ectopic pregnancy}}, highlighting how physical trauma and emotional distress intersect. The lower half of 
Figure~\ref{fig:WC_MentalHealth} 
shows sexual harassment and consent, featuring terms like \textbf{\textit{groping} and \textit{harassed}}, indicating mental-health trauma during such incidents. Under the \textit{\textbf{Pregnancy}} category, the upper section of Figure~\ref{fig:WC_Pregnancy} illustrates terms asking about abortion-related decisions while weighing both good and bad aspects of it. For example, \textbf{\textit{illegal, choice and abortion}}. The middle section of Figure~\ref{fig:WC_Pregnancy} highlights early pregnancy loss indicators and self-monitoring terms such as \textbf{\textit{spotting, discharge, and hCG}} and the lower section reflects clinical follow-up and ultrasound monitoring through terms like \textbf{\textit{fetal pole, gestational sac, and molar}}. Under the \textit{\textbf{Abuse}} category (Figure~\ref{fig:WC_Abuse}), the upper half describes reflective storytelling (\textbf{\textit{telling}, \textit{asking} and \textit{talking}}) and the lower half contains explicit accounts of \textbf{\textit{assault and blackmail.}} Overall, these topics illustrate how users navigate intertwined physical, emotional, and social dimensions of reproductive and trauma-related experiences.

\begin{figure}[!h]
    \centering
    \includegraphics[width=1\linewidth]{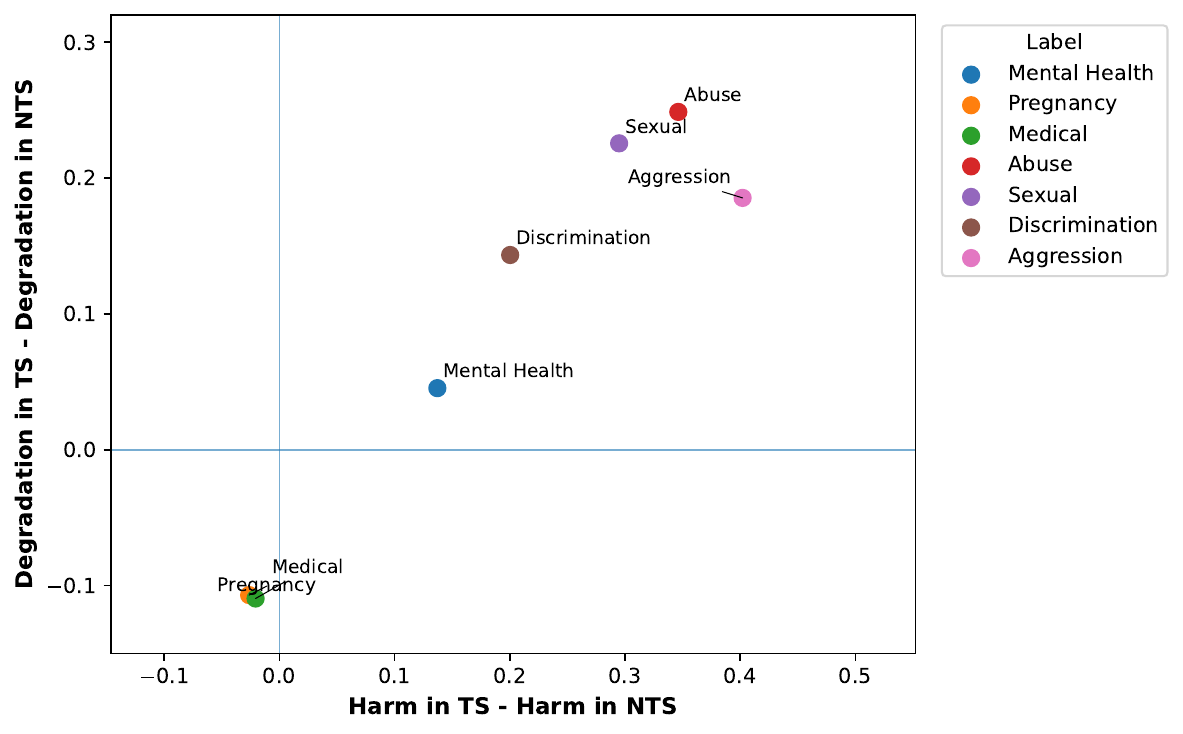}
\caption{With respect to harm and degradation foundations, moral differences between triggering and non-triggering stories. 
    Here, TS stands for Triggering Stories, whereas NTS stands for Non Triggering Stories.
    }
    \label{fig:damf}
\end{figure}

\subsection{Moral Foundations in \corpus}


We identified moral foundations with the Domain Adaptation Moral Foundation (DAMF) model (\citet{guo2023datafusion}). DAMF assigns each story ten scores, one per moral foundation: \textit{care, harm, fairness, cheating, loyalty, betrayal, authority, subversion, purity, and degradation}. We used the highest of these scores to identify the story’s dominant moral foundation. 


Across \corpus, care is the dominant foundation and occurs in 33.30\% of the stories, followed by harm (24.33\%). Authority (13.76\%) and loyalty (13.20\%) form the next tier, while purity (6.77\%) and degradation (5.89\%) appear less often than those mentioned before; fairness (1.86\%), cheating (0.70\%), subversion (0.18\%) and betrayal (0.02\%) rarely dominate. 

Since \textit{Care}, \textit{Authority}, and \textit{Loyalty} are prominent dimensions, we conducted qualitative analysis of the stories identified under them. In these stories, \textit{\textbf{Care}} dimension represents how people show care and concerns about others. For example, ``Should I feel guilty?\ldots I recently spoke out about my perpetrator, and a lot of people found out about it. Now, my perpetrator is showing signs of self-harm\ldots'', illustrates poster (victim) still exhibiting care toward their perpetrator as the latter practices self-harm. In \corpus stories, \textit{\textbf{Authority}} describes how people relate to power, rules, hierarchy, and obedience. ``\ldots The pharmacist refused to fill it\ldots I was denied my medication I needed for my miscarriage\ldots in [location] by a [gender] pharmacist.\ldots'' shows a power imbalance, where the pharmacist acts like a gatekeeper and chooses to block the medicine, which directly affects the patient’s ability to get care for her body.
\textit{\textbf{Loyalty}} is illustrated where posters describe how they expected trust from their partners, which was breached by them in cases of sexual violence.

With respect to harm and degradation foundations, Figure~\ref{fig:damf} compares differences between triggering and non-triggering stories. X-axis shows the differences in average harm scores, whereas y-axis shows differences in average degradation scores between these two groups of stories. Triggering stories pertaining to four categories (\textit{Abuse, Sexual, Aggression, and Discrimination}) show relatively significant differences on both the axes, indicating more harmful and degrading text in these stories. On the other hand, triggering stories in mental health category shows similar degradation levels to the non-triggering ones, but higher harm. However, there is no significant differences found for pregnancy and medical categories.

\subsection{Emotions Expressed in Triggering Stories}

\begin{figure}[!h]
  \centering
  \includegraphics[width=\linewidth]{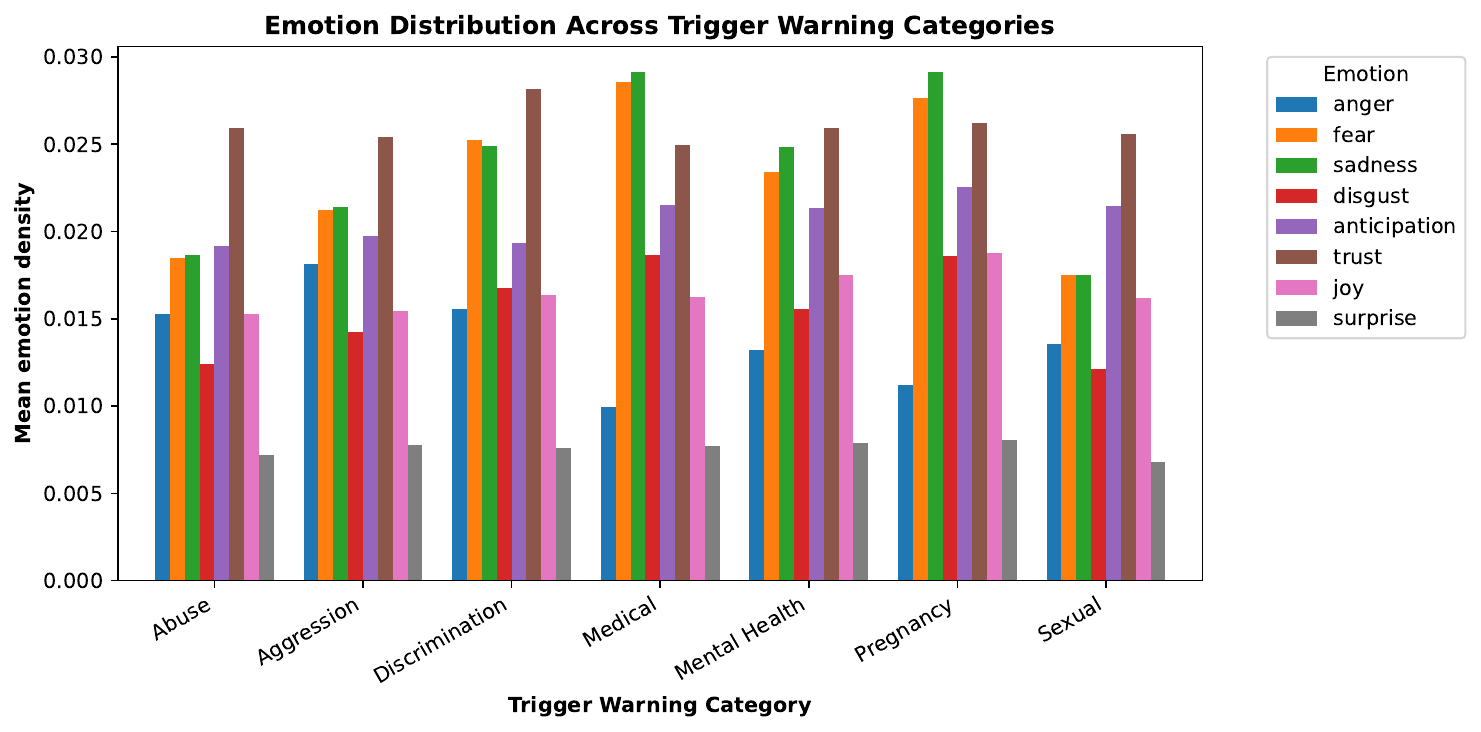}
\caption{Emotion distribution across trigger warning categories based on mean NRC emotion density (anger, fear, sadness, disgust, anticipation, trust, joy, and surprise).}
  \label{fig:psl_emotion}
\end{figure}

Figure~\ref{fig:psl_emotion} presents the distribution of NRC emotion densities across trigger-warning categories. The NRC emotion lexicon operationalizes the eight basic emotions \cite{plutchik2001nature}:  anger, fear, sadness, disgust, anticipation, trust, joy, and surprise. While trust is prevalent across all categories, distinct emotional profiles emerge depending on the underlying topic. Pregnancy and medical-related stories exhibit comparatively higher levels of sadness, consistent with narratives centered on loss, uncertainty, and emotionally challenging bodily experiences. Additionally, stories related to sexual violence, including abuse and sexual categories, show relatively elevated levels of trust-related language. As discussed earlier, trust in this lexicon reflects reliance, expectation, and engagement with interpersonal or institutional frameworks rather than emotional reassurance. Mental health stories display a more balanced distribution across multiple emotions, including sadness, fear, and trust. Overall, these category-level patterns indicate that different trigger warning topics are associated with distinct emotional signatures, even when all categories involve sensitive or distressing content.


Figure~\ref{fig:em_radar} compares the emotions expressed in triggering stories with non-triggering ones (non-triggering stories also include the ones describing topics outside sexual violence and pregnancy-issues). We found that trust as an emotion is prevalent in both types of stories. However, the intensities of other emotions (fear, sadness, disgust, and anticipation) are relatively higher in triggering than in non-triggering stories. For anger, surprise, and joy, there is not much difference being noted.




{\setlength{\fboxrule}{0pt}
\begin{figure}[!h]
    \centering
    \fbox{\includegraphics[width=0.85\linewidth]{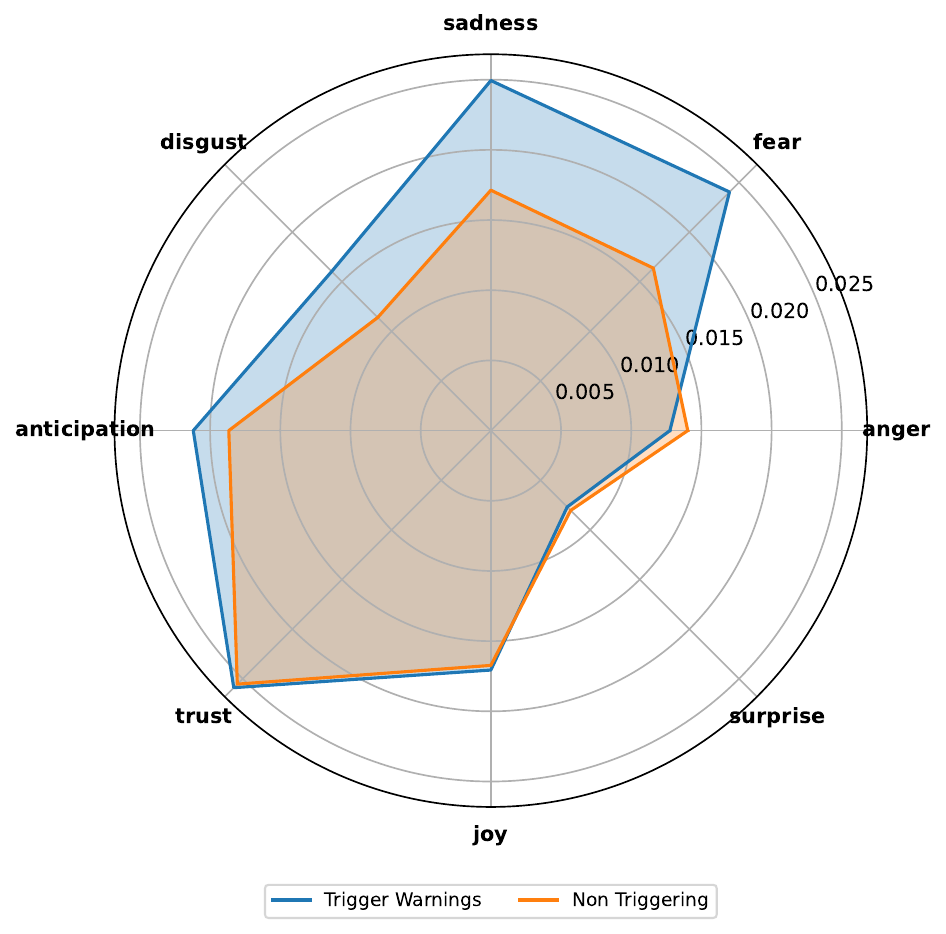}}
\caption{Radar plot of NRC emotion densities comparing triggering stories against non-triggering content. Triggering stories exhibit higher negative emotion density (fear, sadness, disgust, and anticipation), while trust remains prominent across both groups.}
    \label{fig:em_radar}
\end{figure}
}

\section{Related Work}
Prior studies \cite{wiegmann2023trigger, TriggerDataset2023} have approached trigger-warning prediction task as a multi-label classification. However, these works are specific to fanfiction stories. Many studies have also focused on NLP problems pertaining to datasets related to pregnancy and sexual violence. 

\citet{SV-Tracking2020} collect tweets on sexual violence with labeled categories such as type of violence, victim's identity, and gender, and so on. \citet{Safecity2018} gather \np{9892} reports of violence from the SafeCity platform and classified into harassment categories. \citet{garg2025analyzing} release METHREE with 8,947 sentence-level labels from Reddit sexual-violence narratives, and propose sentence-level highlighting of key parts (what happened, effects, and what the author is asking for) to make long stories easier to read. \citet{foriest2024cross} analyze 298 Reddit posts and 10,369 comments across seven subreddits to characterize five forms of muting in gender-based violence discussions and train an ensemble classifier to detect muting using linguistic features. \citet{wyer2025algorithmic} study GPT-3 text completions under gendered prompts and find that outputs about women disproportionately contain sexualized violence, based on topic modeling, sentiment, and toxicity analyses.

\citet{Abortion2023} assemble tweets using selected keywords to capture the full scope of abortion-related discussions in the United States. \citet{AbortionDataset2024} scrape tweets on abortion and miscarriage to gain a deeper understanding of opinions and sentiments. \citet{rao2025polarized} analyze abortion-related tweets and find distinct partisan frames and spikes in hostility around key events. \citet{aleksandric2024analyzing} study post Roe v. Wade leak abortion Facebook stories, train a Transformer stance model, and find stance varies by state and correlates with policy and health indicators. \citet{beel2022linguistic} analyze divisive Reddit discussions (including abortion) and use easy-to-understand text \& user signals (like tone and toxicity) to predict when threads will become heated and why. 

However, none of these studies focuses on annotation based on trigger-warning categories. To the best of our knowledge, this is the first corpus focusing on trigger warnings on women-related issues posted on Reddit.

\section{Conclusion and Discussion}


\textbf{Limitations and Future Work} Our study proposes, \corpus, a dataset of Reddit stories labeled for seven trigger categories. Our work also suffers from a few limitations, each of which leading to future directions. First, our focus is on women-related concerns shared on Reddit. Other topics remain out of scope of this study. Having said that, future works can expand \corpus to additional topics. Moreover, using fine-tuning approaches, knowledge from \corpus can be transferred to new domains. Second, although we experimented with GPT-4o-mini, LLaMA-3-8B, and attempted Gemini (which refused to generate outputs for such sensitive content), future studies may also analyze how other LLMs generate such trigger warnings. Our linguistic analysis revealed annotated stories in \corpus exhibit multiple interesting topics and moral foundations. Hence, it could be beneficial for future researchers in various linguistic ways.

\textbf{FAIR Principles and Ethical Standards} 
\corpus can be accessed through the Zenodo link\footnote{https://doi.org/10.5281/zenodo.18251622} \cite{Tweddit:dataset-Zenodo}. It is accompanied with metadata and unique identifier. It is accessible in CSV format. Standard formats ensure interoperability as well. We removed all Personally Identifiable Information (PII), including usernames, ages, and genders. We even removed the subreddit names to avoid any traceability to actual stories. For code reproducibility, we made our scripts public\footnote{\url{https://github.com/ShirleneRose/TriggerWarnings}}.

This research has been conducted with complete consideration of ethical standards. First, we utilized public stories of Reddit for which, authors' explicit consent is not required \cite{HateIncivility2024}. This is because Reddit has already given consent to make their data available for academic research. Second, because the nature of these stories is disturbing, we did not hire crowd workers for annotating triggering experiences because some survivors among them may get triggered during the annotation task.

\textbf{Implications and Potential Applications} The \corpus dataset provides a unique lens for analyzing sensitive and potentially triggering content in online communities like Reddit. By curating Reddit stories centered around topics like abortion, miscarriage and sexual violence, which are annotated with fine-grained trigger warning categories across domains such as \emph{Mental Health, Abuse, Medical, and Discrimination}. \corpus dataset can enable systematic study of how emotionally charged narratives are framed and interpreted. \corpus's multi-label structure facilitates comparative examination of distinct forms of triggers rather than treating triggering content as a single category.

\corpus further supports the application of advanced natural language processing methods like multi-label classification for automatically generating trigger warnings. \corpus can be leveraged for cross-domain transfer-learning to predict warnings on other platforms such as Gab and Twitter.

Researchers can also uncover emotional signals and latent themes embedded in these sensitive discourses. Beyond academic research, \corpus has practical relevance for informing how online platforms design moderation. By analyzing how trigger warnings are used in community moderated spaces like Reddit, the dataset enables examination of how platforms can protect vulnerable users while still allowing open discussions, highlighting key ethical trade-offs between free expressions and user safety.   
    
\bibliography{aaai2026, references, Vaibhav}

\section{Paper Checklist}

\begin{enumerate}

\item For most authors...
\begin{enumerate}
    \item  Would answering this research question advance science without violating social contracts, such as violating privacy norms, perpetuating unfair profiling, exacerbating the socio-economic divide, or implying disrespect to societies or cultures?
    \answerTODO{Answer} \textcolor{blue}{Yes}

  \item Do your main claims in the abstract and introduction accurately reflect the paper's contributions and scope?
    \answerTODO{Answer} \textcolor{blue}{Yes}
   \item Do you clarify how the proposed methodological approach is appropriate for the claims made? 
    \answerTODO{Answer} \textcolor{blue}{Yes, by testing \cite{wiegmann2023trigger}'s model on \corpus}
   \item Do you clarify what are possible artifacts in the data used, given population-specific distributions?
    \answerTODO{Answer} \textcolor{blue}{Yes}
  \item Did you describe the limitations of your work?
    \answerTODO{Answer} \textcolor{blue}{Yes}
  \item Did you discuss any potential negative societal impacts of your work? \textcolor[HTML]{808080}{NA}
    \answerTODO{Answer}
      \item Did you discuss any potential misuse of your work?
    \answerTODO{Answer} \textcolor[HTML]{808080}{NA}
    \item Did you describe steps taken to prevent or mitigate potential negative outcomes of the research, such as data and model documentation, data anonymization, responsible release, access control, and the reproducibility of findings?
    \answerTODO{Answer} \textcolor{blue}{Yes}
  \item Have you read the ethics review guidelines and ensured that your paper conforms to them?
    \answerTODO{Answer} \textcolor{blue}{Yes}
\end{enumerate}

\item Additionally, if your study involves hypotheses testing...
\begin{enumerate}
  \item Did you clearly state the assumptions underlying all theoretical results?
    \answerTODO{Answer} \textcolor[HTML]{808080}{NA}
  \item Have you provided justifications for all theoretical results?
    \answerTODO{Answer} \textcolor[HTML]{808080}{NA}
  \item Did you discuss competing hypotheses or theories that might challenge or complement your theoretical results?
    \answerTODO{Answer} \textcolor[HTML]{808080}{NA}
  \item Have you considered alternative mechanisms or explanations that might account for the same outcomes observed in your study?
    \answerTODO{Answer} \textcolor[HTML]{808080}{NA}
  \item Did you address potential biases or limitations in your theoretical framework?
    \answerTODO{Answer} \textcolor[HTML]{808080}{NA}
  \item Have you related your theoretical results to the existing literature in social science?
    \answerTODO{Answer} \textcolor[HTML]{808080}{NA}
  \item Did you discuss the implications of your theoretical results for policy, practice, or further research in the social science domain?
    \answerTODO{Answer} \textcolor[HTML]{808080}{NA}
\end{enumerate}

\item Additionally, if you are including theoretical proofs...
\begin{enumerate}
  \item Did you state the full set of assumptions of all theoretical results?
    \answerTODO{Answer} \textcolor[HTML]{808080}{NA}
	\item Did you include complete proofs of all theoretical results?
    \answerTODO{Answer} \textcolor[HTML]{808080}{NA}
\end{enumerate}

\item Additionally, if you ran machine learning experiments...
\begin{enumerate}
  \item Did you include the code, data, and instructions needed to reproduce the main experimental results (either in the supplemental material or as a URL)?
    \answerTODO{Answer} \textcolor{blue}{Yes, dataset and code reproducibility links are provided.}
  \item Did you specify all the training details (e.g., data splits, hyperparameters, how they were chosen)?
    \answerTODO{Answer} \textcolor{blue}{Yes, training settings and parameters are clearly documented.}

     \item Did you report error bars (e.g., with respect to the random seed after running experiments multiple times)?
    \answerTODO{Answer} \textcolor[HTML]{808080}{NA}
	\item Did you include the total amount of compute and the type of resources used (e.g., type of GPUs, internal cluster, or cloud provider)?
    \answerTODO{Answer} \textcolor[HTML]{808080}{NA}
     \item Do you justify how the proposed evaluation is sufficient and appropriate to the claims made? 
    \answerTODO{Answer} \textcolor[HTML]{808080}{NA}
     \item Do you discuss what is ``the cost`` of misclassification and fault (in)tolerance?
    \answerTODO{Answer} \textcolor[HTML]{808080}{NA}
  
\end{enumerate}

\item Additionally, if you are using existing assets (e.g., code, data, models) or curating/releasing new assets, \textbf{without compromising anonymity}...
\begin{enumerate}
  \item If your work uses existing assets, did you cite the creators?
    \answerTODO{Answer} \textcolor[HTML]{808080}{NA} 
  \item Did you mention the license of the assets?
    \answerTODO{Answer} \textcolor[HTML]{808080}{NA}
  \item Did you include any new assets in the supplemental material or as a URL?
    \answerTODO{Answer} \textcolor{blue}{Yes}
  \item Did you discuss whether and how consent was obtained from people whose data you're using/curating?
    \answerTODO{Answer} \textcolor{blue}{Yes}
  \item Did you discuss whether the data you are using/curating contains personally identifiable information or offensive content?
    \answerTODO{Answer} \textcolor{blue}{Yes}
\item If you are curating or releasing new datasets, did you discuss how you intend to make your datasets FAIR (see \citet{fair})?
\answerTODO{Answer} \textcolor{blue}{Yes}
\item If you are curating or releasing new datasets, did you create a Datasheet for the Dataset (see \citet{gebru2021datasheets})? 
\answerTODO{Answer} \textcolor{blue}{Yes}
\end{enumerate}

\item Additionally, if you used crowdsourcing or conducted research with human subjects, \textbf{without compromising anonymity}...
\begin{enumerate}
  \item Did you include the full text of instructions given to participants and screenshots?
    \answerTODO{Answer} \textcolor[HTML]{808080}{NA} 
  \item Did you describe any potential participant risks, with mentions of Institutional Review Board (IRB) approvals?
    \answerTODO{Answer} \textcolor[HTML]{808080}{NA} 
  \item Did you include the estimated hourly wage paid to participants and the total amount spent on participant compensation?
    \answerTODO{Answer} \textcolor[HTML]{808080}{NA} 
   \item Did you discuss how data is stored, shared, and deidentified?
   \answerTODO{Answer} \textcolor[HTML]{808080}{NA} 
\end{enumerate}

\end{enumerate}

\clearpage
\section{Appendix}
\label{sec:appendix}

We use output perplexity (PPL) as an uncertainty signal to identify predictions likely to be incorrect.
For each story, we record the maximum perplexity score across the seven trigger categories and flag stories
above a chosen threshold for manual review. We evaluate a range of thresholds and report precision, recall, F1
trade-offs for identifying mispredicted instances.

\FloatBarrier

\begin{table}[H]
\centering
\caption{Round 1 threshold sweep for perplexity-based review.}
\label{tab:ppl_threshold_r1}
\small
\begin{tabular}{lrrrrrrr}
\toprule
Threshold & TP & FP & FN & TN & Precision & Recall & F1 \\
\midrule
1.0000 & 62 & 43 & 0 & 0 & 0.5905 & 1.0000 & 0.7425 \\
1.0005 & 60 & 30 & 2 & 13 & 0.6667 & 0.9677 & 0.7895 \\
\textbf{1.0007} & \textbf{60} & \textbf{30} & \textbf{2} & \textbf{13} & \textbf{0.6667} & \textbf{0.9677} & \textbf{0.7895} \\
1.0010 & 59 & 29 & 3 & 14 & 0.6705 & 0.9516 & 0.7867 \\
1.0020 & 57 & 28 & 5 & 15 & 0.6706 & 0.9194 & 0.7755 \\
\bottomrule
\end{tabular}
\end{table}

\begin{table}[H]
\centering
\caption{Round 2 threshold sweep for perplexity-based review.}
\label{tab:ppl_threshold_r2}
\small
\begin{tabular}{lrrrrrrr}
\toprule
Threshold & TP & FP & FN & TN & Precision & Recall & F1 \\
\midrule
1.0000 & 47 & 43 & 0 & 0 & 0.5222 & 1.0000 & 0.6861 \\
1.0001 & 46 & 37 & 1 & 6 & 0.5542 & 0.9787 & 0.7077 \\
1.0002 & 46 & 36 & 1 & 7 & 0.5610 & 0.9787 & 0.7132 \\
\textbf{1.0007} & \textbf{43} & \textbf{35} & \textbf{4} & \textbf{8} & \textbf{0.5513} & \textbf{0.9149} & \textbf{0.6880} \\
1.0020 & 39 & 34 & 8 & 9 & 0.5342 & 0.8298 & 0.6500 \\
1.0070 & 37 & 31 & 10 & 12 & 0.5441 & 0.7872 & 0.6435 \\
\bottomrule
\end{tabular}
\end{table}

\begin{table}[H]
\centering
\caption{Round 3 threshold sweep for perplexity-based review.}
\label{tab:ppl_threshold_r3}
\small
\begin{tabular}{lrrrrrrr}
\toprule
Threshold & TP & FP & FN & TN & Precision & Recall & F1 \\
\midrule
1.0000 & 48 & 41 & 9 & 2 & 0.5393 & 0.8421 & 0.6575 \\
1.0001 & 47 & 36 & 10 & 7 & 0.5663 & 0.8246 & 0.6714 \\
1.0003 & 47 & 33 & 10 & 10 & 0.5875 & 0.8246 & 0.6861 \\
\textbf{1.0007} & \textbf{44} & \textbf{32} & \textbf{13} & \textbf{11} & \textbf{0.5789} & \textbf{0.7719} & \textbf{0.6617} \\
1.0030 & 43 & 26 & 14 & 17 & 0.6232 & 0.7544 & 0.6825 \\
1.1738 & 27 & 7 & 30 & 36 & 0.7941 & 0.4737 & 0.5934 \\
\bottomrule
\end{tabular}
\end{table}

\begin{table}[H]
\centering
\caption{Round 4 threshold sweep for perplexity-based review.}
\label{tab:ppl_threshold_r4}
\small
\begin{tabular}{lrrrrrrr}
\toprule
Threshold & TP & FP & FN & TN & Precision & Recall & F1 \\
\midrule
1.0000 & 44 & 41 & 0 & 0 & 0.5176 & 1.0000 & 0.6822 \\
1.0001 & 40 & 35 & 4 & 6 & 0.5333 & 0.9091 & 0.6723 \\
1.0004 & 37 & 31 & 7 & 10 & 0.5441 & 0.8409 & 0.6607 \\
\textbf{1.0007} & \textbf{36} & \textbf{29} & \textbf{8} & \textbf{12} & \textbf{0.5538} & \textbf{0.8182} & \textbf{0.6606} \\
1.0067 & 31 & 22 & 13 & 19 & 0.5849 & 0.7045 & 0.6392 \\
1.0070 & 30 & 22 & 14 & 19 & 0.5769 & 0.6818 & 0.6250 \\
\bottomrule
\end{tabular}
\end{table}
\FloatBarrier

\FloatBarrier
\begin{figure}[H]
    \centering
    \begin{subfigure}[t]{0.48\linewidth}
        \centering
        \includegraphics[width=\linewidth]{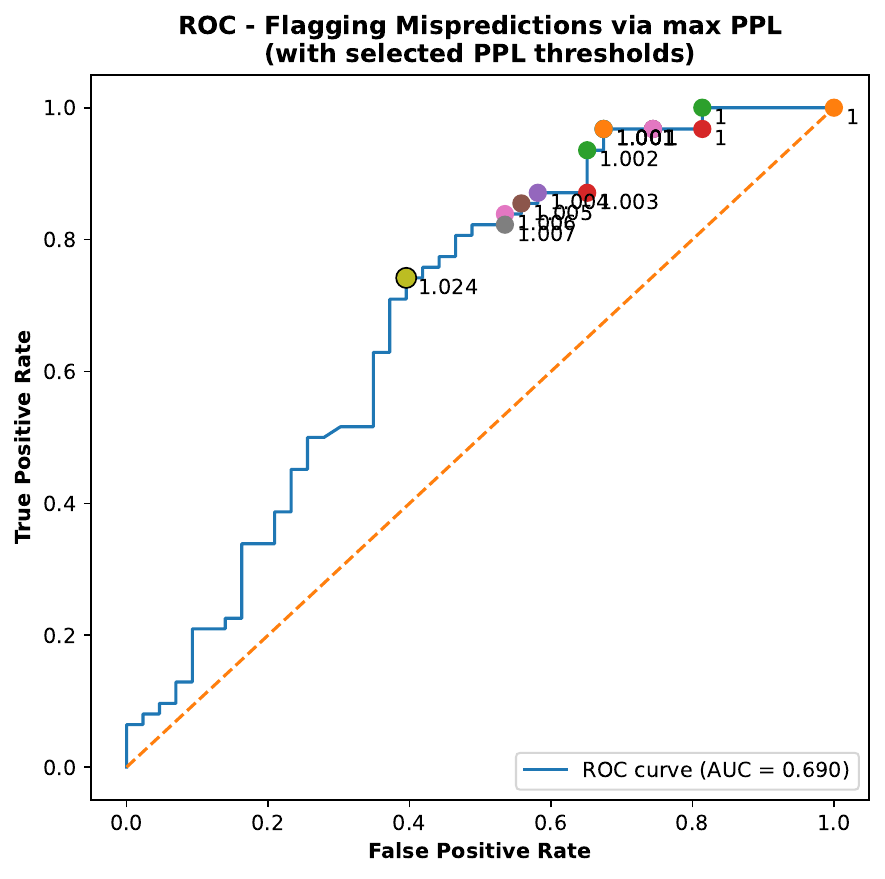}
        \caption{Round 1 ROC curve.}
    \end{subfigure}
    \hfill
    \begin{subfigure}[t]{0.48\linewidth}
        \centering
        \includegraphics[width=\linewidth]{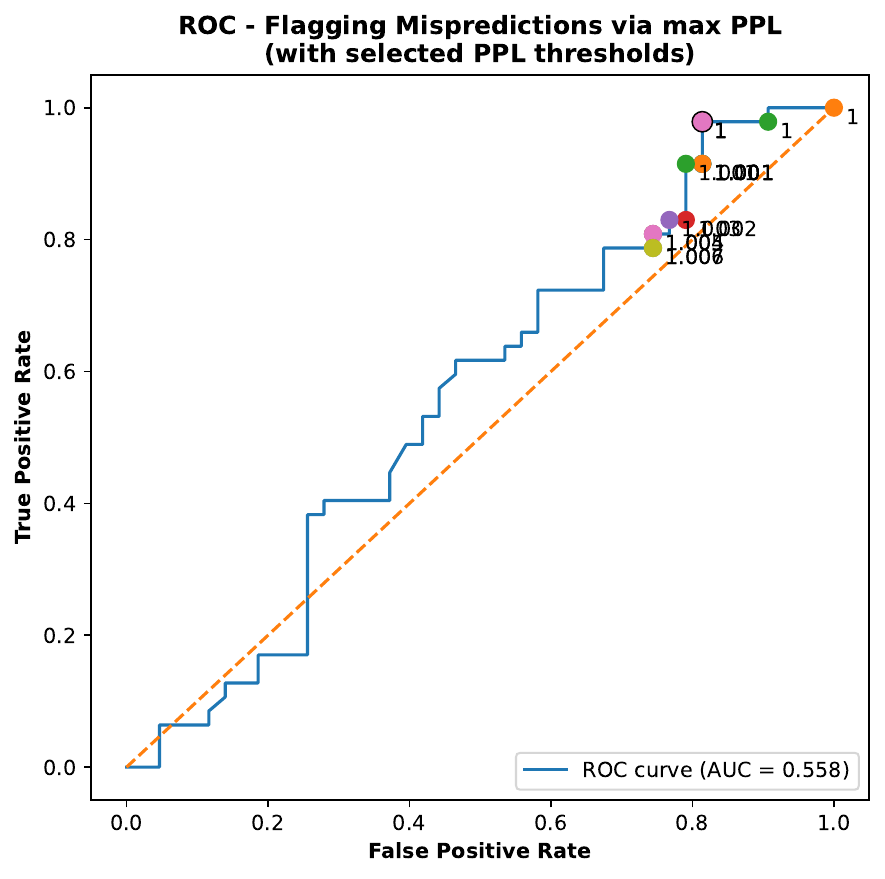}
        \caption{Round 2 ROC curve.}
    \end{subfigure}

    \vspace{0.8em}

    \begin{subfigure}[t]{0.48\linewidth}
        \centering
        \includegraphics[width=\linewidth]{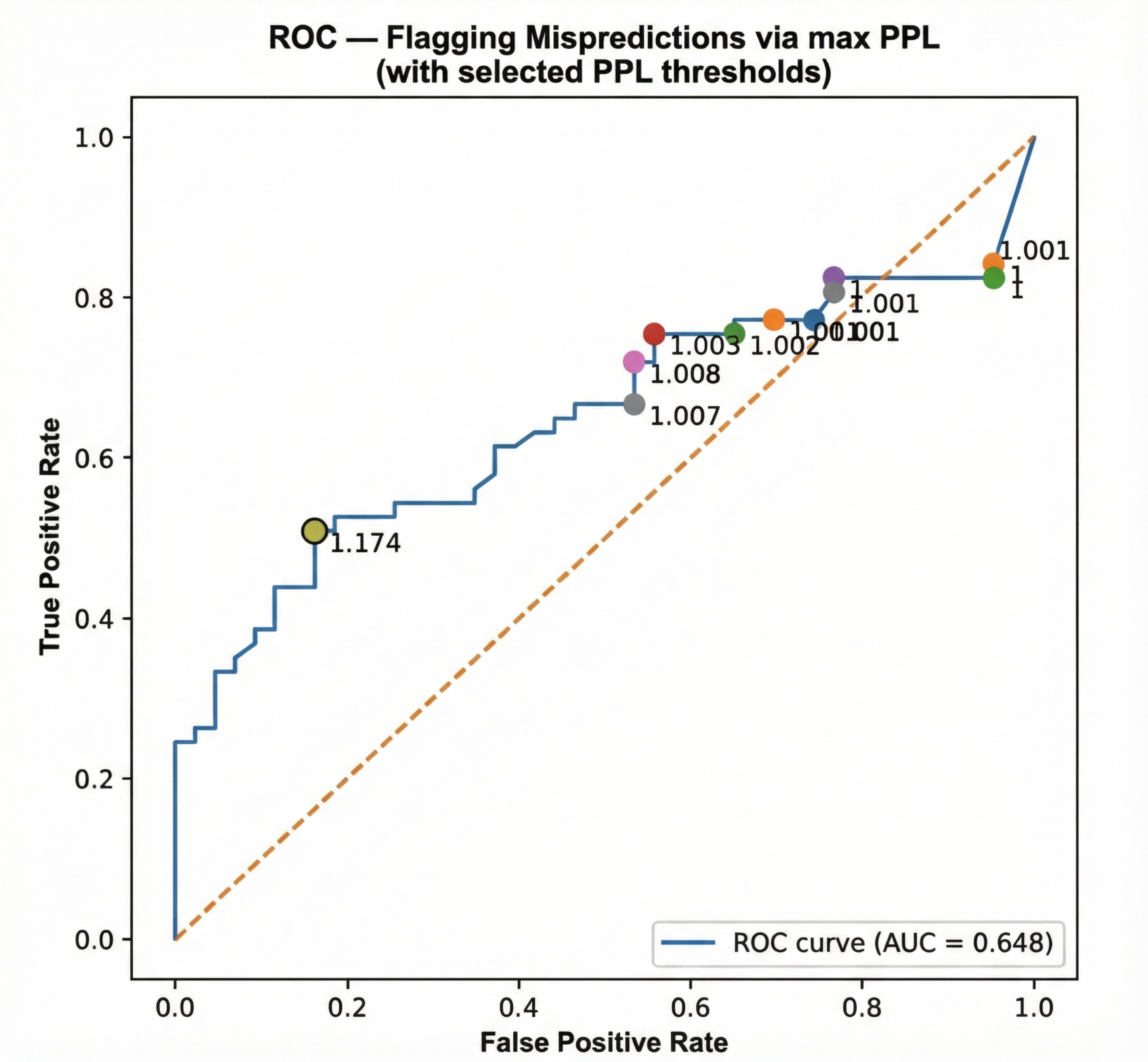}
        \caption{Round 3 ROC curve.}
    \end{subfigure}
    \hfill
    \begin{subfigure}[t]{0.48\linewidth}
        \centering
        \includegraphics[width=\linewidth]{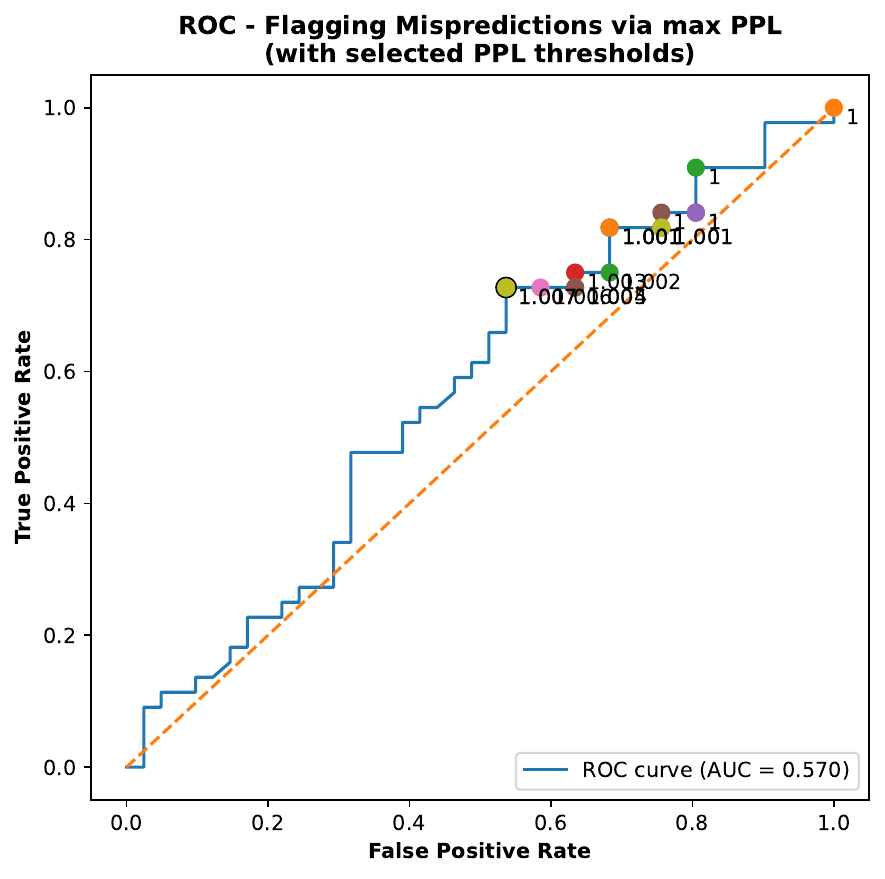}
        \caption{Round 4 ROC curve.}
    \end{subfigure}

    \caption{ROC curves for identifying mispredicted stories using maximum output perplexity (max PPL) as an uncertainty score across trigger categories. Higher max PPL indicates lower model confidence, and thresholds along the curve correspond to different precision--recall trade-offs for selecting stories for manual review.}
    \label{fig:roc_all_rounds}
\end{figure}

\end{document}